\documentclass[conference]{IEEEtran}
\IEEEoverridecommandlockouts
\usepackage{cite}
\usepackage{amsmath,amssymb,amsfonts}
\usepackage{graphicx}
\usepackage{textcomp}
\usepackage{xcolor}

\usepackage[utf8]{inputenc} % allow utf-8 input
\usepackage[T1]{fontenc}    % use 8-bit T1 fonts
\usepackage{hyperref}       % hyperlinks
\usepackage{url}            % simple URL typesetting
\usepackage{booktabs}       % professional-quality tables
\usepackage{amsfonts}       % blackboard math symbols
\usepackage{nicefrac}       % compact symbols for 1/2, etc.
\usepackage{microtype}      % microtypography
\usepackage{xcolor}         % colors
\usepackage{graphicx} % for including images
\usepackage{caption}
\usepackage{subcaption}
\usepackage{algpseudocode}
\usepackage{amsmath} % For advanced math typesetting
\usepackage{booktabs}
\usepackage[export]{adjustbox}

\def\BibTeX{{\rm B\kern-.05em{\sc i\kern-.025em b}\kern-.08em
    T\kern-.1667em\lower.7ex\hbox{E}\kern-.125emX}}
\begin{document}

\title{Graph-Based Biomarker Discovery and Interpretation for Alzheimer's Disease}

\author{\IEEEauthorblockN{Maryam Khalid
\IEEEauthorrefmark{1}\thanks{MK and FSK have equal contribution.},
Fadeel Sher Khan\IEEEauthorrefmark{2},
John~Broussard\IEEEauthorrefmark{3},
Arko~Barman\IEEEauthorrefmark{1}}
\IEEEauthorblockA{\IEEEauthorrefmark{1}Rice University}
\IEEEauthorblockA{\IEEEauthorrefmark{2}University of Texas at Austin}
\IEEEauthorblockA{\IEEEauthorrefmark{3}University of Texas Health Science Center at Houston\\
Email: arko.barman@rice.edu}}

% \author{\IEEEauthorblockN{****** ******\IEEEauthorrefmark{1}\thanks{** and *** have equal contribution.},
% ****** **** ****\IEEEauthorrefmark{2},
% **** *********\IEEEauthorrefmark{3},
% **** ******\IEEEauthorrefmark{1}}
% \IEEEauthorblockA{\IEEEauthorrefmark{1}*******}
% \IEEEauthorblockA{\IEEEauthorrefmark{2}*******}
% \IEEEauthorblockA{\IEEEauthorrefmark{3}*******\\
% Email: *********}}

\maketitle

\begin{abstract}
Early diagnosis and discovery of therapeutic drug targets are crucial objectives for effective management of Alzheimer’s Disease (AD). Current approaches for AD diagnosis and treatment planning are based on radiological imaging and largely inaccessible for population-level screening due to prohibitive costs and limited availability. Recently, blood tests have shown promise in diagnosing AD and highlighting possible biomarkers that can be used as drug targets for AD management. Blood tests are significantly more accessible to disadvantaged populations, cost-effective, and minimally invasive. However, biomarker discovery in the context of AD diagnosis is complex as there exist important associations between various biomarkers. Here, we introduce BRAIN (Biomarker Representation, Analysis, and Interpretation Network), a novel machine learning (ML) framework to jointly optimize diagnostic accuracy and biomarker discovery processes to identify all relevant biomarkers that contribute to AD diagnosis. Using a holistic graph-based representation for biomarkers, we highlight their interdependencies and explain why different ML models identify different discriminative biomarkers. We apply BRAIN to a publicly available blood biomarker dataset, revealing three novel biomarker subnetworks whose interactions vary between the control and AD groups, offering a new paradigm for drug discovery and biomarker analysis for AD.
\end{abstract}


\section{Introduction}
Currently, more than 55 million people live with dementia globally, with 10 million new cases per year. Dementia is the 7th leading cause of death and one of the leading causes of disability and dependency among older people worldwide. Alzheimer's Disease (AD), the most common form of dementia, makes up 60-70\% of total cases~\cite{o}. AD reduces lifespan by 8-13 years for those diagnosed in their 60s or 70s~\cite{zanetti}. The current standard of care for AD is the use of monoclonal antibodies such as lecanamab or donanemab~\cite{bentley_monoclonal_2024, sims_donanemab_2023}. Recent findings indicate that the efficacy of these therapies is greater during earlier stages of AD, making early diagnosis crucial for treatment~\cite{mintun_donanemab_2021}. Early diagnosis can help promote the use of therapeutic treatment to reduce amyloid or tau burden and inflammation contributing to neurodegeneration~\cite{bentley_multimodal_nodate}. Early diagnosis can also allow identification and treatment accompanying physical illnesses and risk factors, thus optimizing physical health, cognition, activity, and wellbeing \cite{o}.

The current gold standard for AD diagnosis is medical structural imaging, which focuses on the identification of abnormal formations inside the brain and thus has a significant risk of missing early AD diagnosis~\cite{l21}. Moreover, these methods do not capture possible risk factors or potential drug targets associated with AD. This is potentially harmful as research evidence points to racial disparities in AD diagnosis and identifying biological risk factors as intrinsically linked with early AD diagnosis for effective management of these disease conditions~\cite{ahs}. Hence, identifying blood-based protein biomarkers for AD is increasingly seen as a promising avenue to accelerate early diagnosis and treatment of AD. Blood tests offer a cost-effective, minimally invasive, and easily accessible solution for AD detection, and can detect molecular changes indicative of AD pathology, including abnormal levels of different biomarkers, even before clinical symptoms manifest. Biomarkers can provide insight into the biochemical processes contributing to AD pathogenesis~\cite{Hampel2023}. This knowledge is invaluable for the development of targeted therapies. The combination of various biomarkers might allow the identification of disease subtypes and personalized treatment approaches, addressing the heterogeneity of AD more effectively than current methods.

\begin{figure*}[t]
    \centering
    \includegraphics[width = .6\textwidth]{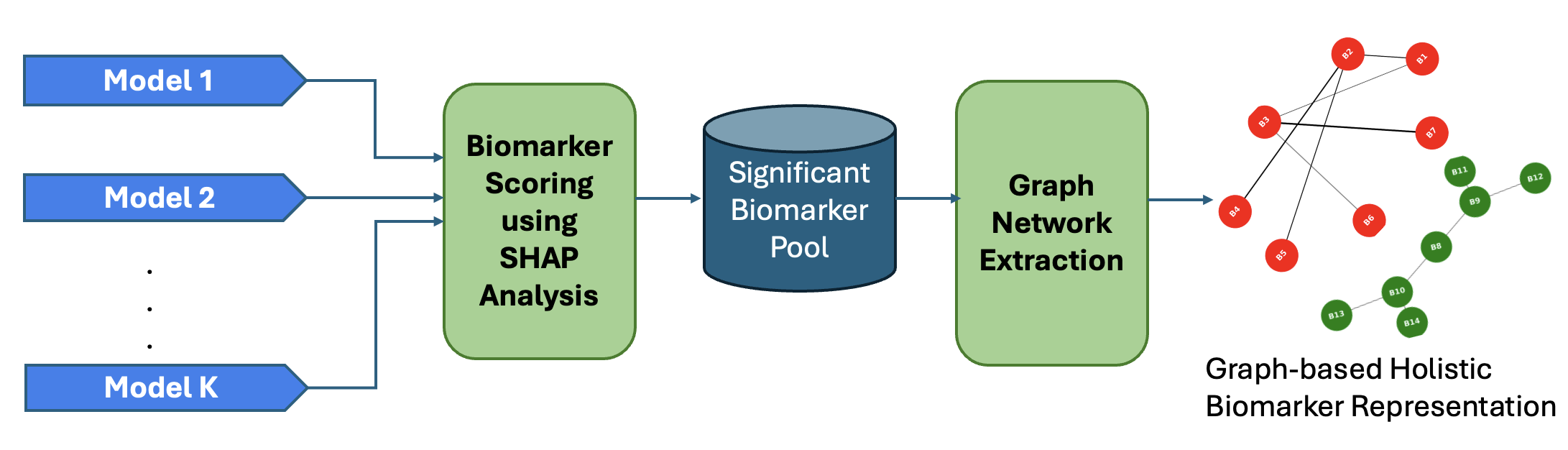}
    \caption{BRAIN: Biomarker Exploration and Representation Network}
    \label{fig:framework}
\end{figure*}

To this end, we systematically designed BRAIN: \textbf{\underline{B}}iomarker \textbf{\underline{R}}epresentation, \underline{\textbf{A}}nalysis and \textbf{\underline{I}}nterpretation \textbf{\underline{N}}etwork that provides holistic insights into biomarker discovery from multiple machine learning (ML) models. It is tailored to robustly handle the complexity of biomarker identification considering a diverse range of $K$ ML models, each optimizing different objectives and varying in complexity. As shown in Figure \ref{fig:framework}, the process begins with training these multiple ML models in a bootstrapped fashion, enhancing robustness. Once trained, importance scores are assigned to features using SHapley Additive exPlanations (SHAP) analysis~\cite{shap}. In the next step, BRAIN aggregates these importance scores assigned to each biomarker across the ensemble of models. This aggregation helps identify a wide pool of critical biomarkers, encompassing those deemed significant across diverse models \cite{Hampel2023-personalizedmed}. 

Subsequently, BRAIN analyzes this pool of critical biomarkers. As the size of important biomarker sets increases, interpretability and explainability become significant challenges, which are further exacerbated when the relationships between these biomarkers are also considered. To address these issues, BRAIN extracts novel graph representations for these biomarkers, enhancing interpretability. This representation encapsulates the relationships and interactions between the identified biomarkers, offering a visual depiction of their interconnectedness. By leveraging this graph network representation, BRAIN facilitates the exploration and interpretation of biomarker relationships, aiding researchers gaining deeper insights into the underlying biology and pathology.

We summarize the key contributions of our work as follows:
\begin{itemize}
    \item We present a framework to identify a comprehensive set of important AD biomarkers by utilizing diverse ML models, enhancing robustness and mitigating model bias.

    \item We present novel interpretable biomarker networks for AD, delineating three distinct clusters and showcasing multiple distinguishing network aspects between AD and control graphs. We highlight interdependencies between biomarkers and how they change between normal and AD, providing new biomedical insights.

\end{itemize}

While BRAIN can be applied to any dataset with blood biomarkers for both AD and controls, we focus our analysis on the open access Texas Alzheimer's Research and Care Consortium (TARCC) dataset consisting of blood biomarker data from AD patients and controls~\cite{OBryant2008}. We discover three sub-networks that exist in biomarker interactions in the dataset and use graph theory to explain how they are different between the AD and control groups. A comparative examination of these clusters reveals distinct differences in network topology, indicating substantial alterations in network-level interactions between control and AD patient populations. Our findings suggest that when viewed in isolation, individual biomarkers are insufficient to capture the complexity of AD pathology, but can offer biomedically-meaningful insights when presented as part of a holistic network.

\subsection{Blood Biomarkers for Alzheimer's Disease Detection}

The clinical potential of blood-based indicators associated with AD, namely amyloid \(\beta\) (A\(\beta\)1-42 and A\(\beta\)1-40), phosphorylated tau (pTau), neurofilament light chain (NfL), and glial fibrillary acidic protein (GFAP), are significant as they show strong associations with those indicators used by standard medical imaging methods~\cite{Blennow2018}. Despite these advancements, it is challenging to achieve a holistic understanding of how these biomarkers interconnect and correlate with each other and with the multifaceted pathophysiology of AD~\cite{Hampel2023-personalizedmed}. This comprehensive understanding is crucial to optimize the diagnostic and prognostic utility of blood-based biomarkers and to pave the way to personalized medicine in AD. Addressing this challenge requires concerted efforts in research to explain the complex interactions among these biomarkers and to validate their collective diagnostic value across diverse populations.

A common approach in biomarker identification is the panel-based method, which involves generating multiple sets of biomarkers, training predictive models, and selecting the set with the highest accuracy metric. The primary objective of these studies is solely to maximize accuracy, leading them to eliminate a large set of biomarkers before training the model. This could potentially eliminate redundant but important biomarkers, thus affecting the discovery process. At the same time, a key limitation of these methods is that they are driven towards finding the \textit{smallest subset} of biomarkers that maximize model performance, potentially overlooking many important biomarkers. These works often fail to explore the interdependency between biomarkers, instead observing them in isolation purely as predictive features.

Various studies in the literature employing similar models have unearthed disparate biomarkers, sometimes even when utilizing identical datasets \cite{s19}.  This is attributed to the different objective functions being optimized. For instance, logistic regression optimizes the likelihood of an observation, while tree-based models optimize the splits on each node based on an impurity metric. The implementation of the model and the interplay between different features also influence which biomarkers are deemed significant for AD. While accuracy metrics aid in identifying the most effective model, determining the optimal model in terms of insights into disease and potential biomarker drug targets remains an open question \cite{review}. Models achieving high accuracy may focus on a select few biomarkers, potentially overlooking a spectrum of other important ones. Addressing this discrepancy and investigating why similar models, even when applied to the same datasets, yield distinct biomarkers poses significant challenges.

\begin{figure*}[ht]
    \centering
    \subfloat[][\centering Age Distribution]{{\includegraphics[width=0.45\textwidth ,height=4.5cm, keepaspectratio]{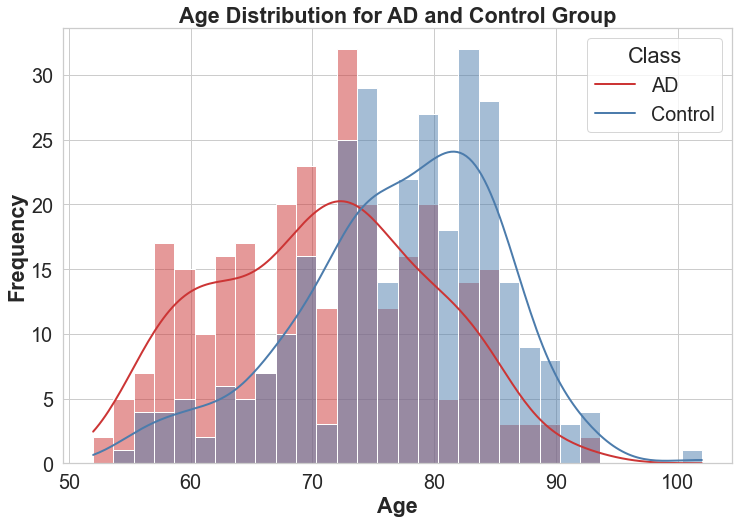} }}%
    \hspace{1cm} 
    \subfloat[][\centering Race Distribution]{{\includegraphics[width=0.45\textwidth,height=4.5cm, keepaspectratio]{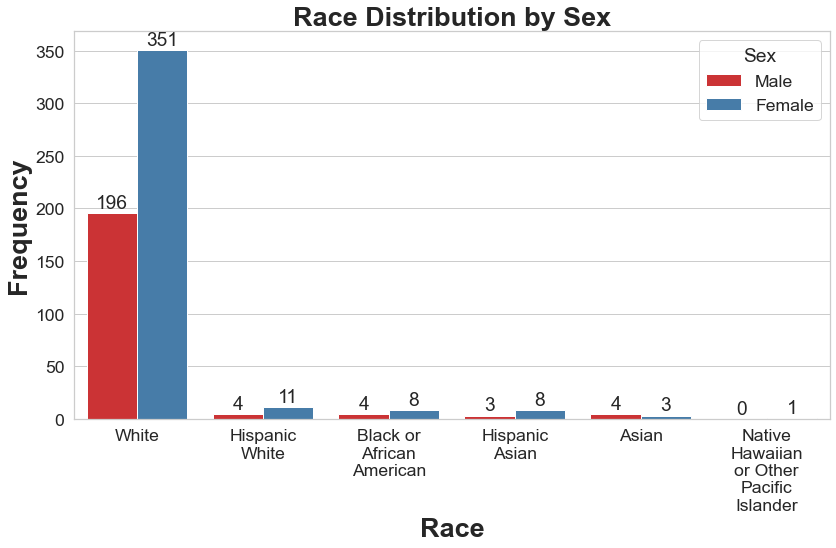} }}%
    \caption{Dataset Demographics}%
    \label{fig:demographics}%
\end{figure*}

Compared to previous works, our objective is to optimize both detection accuracy and biomarker discovery while enhancing interpretability. %, we develop an approach with less need for rigid feature selection.  
Our approach aims to (a) find the comprehensive set of biomarkers that maximize the discriminability of AD and control groups across diverse ML models (Section \ref{brain-intro}), and (b) present an interpretable representation of the interconnectedness and association of these biomarkers and how it varies between AD and control groups (Section \ref{graphs-section}). 

\section{Related Work}
As blood tests provide a minimally invasive, widely accessible and cost-effective solution for AD detection, some ML models have been developed to detect AD from blood biomarkers. While sophisticated models have been devised for AD detection leveraging expensive medical imaging techniques (MRI, EEG, and PET scans), the exploration of cost-effective biomarkers has received limited attention~\cite{review}. In most related work, sophisticated deep learning (DL) models are seldom used for blood biomarkers, primarily due to the scarcity of data.

Existing methods focus on exploration of different combinations of biomarkers to maximize predictive performance. Support Vector Machines (SVM)~\cite{svm2,svm3,svm4} and logistic regression (LR)\cite{logistic1}\cite{logistic2} are frequently employed ML models in AD detection. Random Forest (RF)~\cite{rf1,rf2} is another popular choice along with Naïve Bayes~\cite{naive} to identify a panel of important biomarkers. Limited works that use DL include~\cite{ann} and~\cite{nn}. AutoML technology Just Add Data Bio (JADBIO) was also to train multiple ML models and choose the most suitable model~\cite{svm}. The study focused on AD datasets that suffered from curse of dimensionality due to small sample size. The analysis identified Ridge LR, SVM and RF models to provide optimal performance for protein, miRNA and mRNA features respectively. Important biomarkers have been identified using statistically equivalent signature (SES) algorithm \cite{stat_FI}.

While these approaches focus primarily on maximizing predictive accuracy, they provide limited insight into the relationships and interdependencies among biomarkers. Prior work has explored interpretable graph models via predefined biological hierarchies (e.g., NeuroTree~\cite{neurotree}), sparse network estimation techniques such as graphical lasso~\cite{graph_lasso}, and GNN explainability methods (e.g., GNNExplainer~\cite{gnnexplainer}). These approaches either assume known graph structure, focus on unsupervised conditional dependencies, or explain predictions of trained graph models. In contrast, our work addresses a complementary problem: discovering disease-specific, interpretable biomarker interaction networks directly from tabular clinical data without assuming prior pathway structure, using SHAP-guided feature attribution to ground network construction in predictive relevance. By comparing a shallow neural network with RF and LR models, we emphasize the importance of exploring alternative approaches that could facilitate more widespread and cost-effective screening for AD while providing clinically meaningful insights into biomarker relationships.

\section{Data}

We utilize publicly available data from the Texas Harris Alzheimer’s Research Study by the Texas Alzheimer's Research and Care Consortium (TARCC)~\cite{OBryant2008}. Although several AD cohorts are publicly available, many prioritize imaging or CSF biomarkers or provide blood biomarker data only for subsets of participants with limited harmonization across visits. Therefore, we focus on TARCC for its relatively comprehensive and longitudinal blood biomarker panel, while recognizing that external validation on additional cohorts will be important as comparable datasets become available. The dataset includes anonymized records of patient visits from 2007-2016, gathered across seven clinics. It encompasses 14,655 clinical visit data points across 3,670 unique patients, with 943 variables each. Every visit is treated as an independent sample for analysis, with the physician’s diagnosis acting as the target outcome. The possible diagnoses are AD, Mild Cognitive Impairment (MCI), or Cognitively Normal (CN). Due to the class imbalance with MCI and the complexity of resampling in a dataset with a large number of features, this study specifically focuses on patients diagnosed as either CN or control (N=1555) or AD (N=1320). While disease progression over time is acknowledged (e.g., a shift from CN to MCI or AD in subsequent visits), such cases are considered outside the scope of this analysis, and each visit is analyzed as a separate entity, although age is controlled for in our experimental setup. Additionally, only data relating to blood biomarkers is included in this investigation. For additional information on data variables and access requests, refer to TARCC~\cite{OBryant2008}. The distributions of race and age are provided in Figure~\ref{fig:demographics}. The age distribution, spanning from 50 to 100 years, exhibits distinct skewness towards higher age brackets, with a discernible shift in mean values between AD and control groups. The dataset is highly concentrated around White participants with very few samples from underrepresented groups. Moreover, there are close to twice as many male participants compared to female participants. This points to racial disparities in AD studies. In our work, we only utilized blood biomarker features consisting of proteomics data, i.e., concentration levels of proteins in a blood sample. The blood biomarker subset consists of 593 patient samples, each with 195 biomarkers.

\section{BRAIN: Biomarker Exploration and Representation Network Framework} \label{brain-intro}
\subsection{Exploratory analysis for biomarker discriminability}
\begin{figure}[t]
    \centering
    \includegraphics[trim ={5 0 0 28}, clip, width=.7\linewidth]{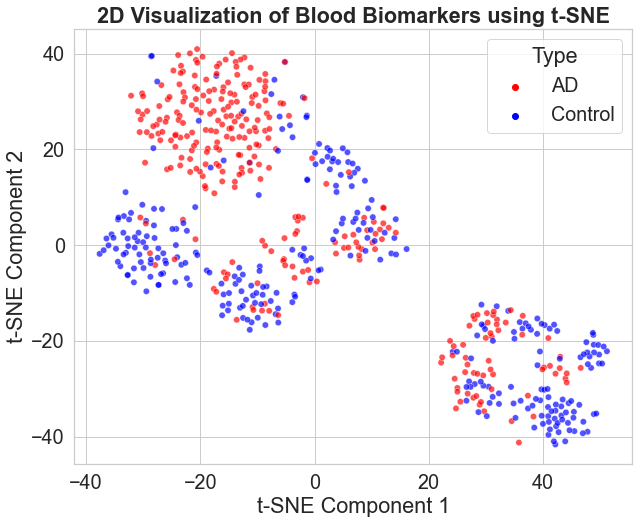}
    \caption{t-SNE Analysis of blood biomarkers highlights distinct AD clusters}
    \label{fig:tsne}
\end{figure}

To demonstrate the discriminative power of the biomarker dataset in the classification of AD vs. control, feature selection was carried out using maximum-relevance-minimum-redundancy (mRMR)~\cite{dp03} and visualized using t-Distributed Stochastic Neighbor Embedding (t-SNE) (see Figure \ref{fig:tsne}). 

mRMR is a filter-based feature selection method that selects the most informative features while minimizing redundancy between the selected features by first finding redundancy with the least correlated features, e.g., through minimizing pairwise correlations. Thereafter, features are selected one by one by applying a greedy search to maximize the objective function, which is a function of relevance and redundancy. This was implemented by first applying a function for mutual information, \textit{I(x,y)}, based on marginal probabilities for \textit{x, y}:

\begin{equation}
\label{eq1}
I(x, y) = \sum_{i,j} p(x_i, y_j) \log \frac{p(x_i, y_j)}{p(x_i)p(y_j)}
\end{equation}

Redundancy ($W_I$) and relevancy ($V_I$) were calculated as,

% \begin{equation}
% \label{eq2}
% W_I = \frac{1}{|S|^2} \sum_{i,j \in S} I(i, j)
% \end{equation}

% \begin{equation}
% \label{eq3}
% V_I = \frac{1}{|S|} \sum_{i \in S} I(h, i)
% \end{equation}

% \begin{minipage}{0.46\textwidth}
\begin{equation}
\label{eq2}
W_I = \frac{1}{|S|^2} \sum_{i,j \in S} I(i, j)
\end{equation}
% \end{minipage}
% \hfill % Fill the space between minipages
% \begin{minipage}{0.48\textwidth}
\begin{equation}
\label{eq3}
V_I = \frac{1}{|S|} \sum_{i \in S} I(h, i)
\end{equation}
% \end{minipage}
where \textit{S} is the feature set, \textit{h} is the response variable, and \textit{i} and \textit{j} are distinct features. Then, the greedy search optimization algorithm searches all combinations of \textit{S} to determine the best features by maximizing $V_I/W_I$.

After selecting the best features, t-SNE was employed to visualize the capability of blood biomarkers in distinguishing between AD and control (CN) cases. t-SNE is a non-linear dimensionality reduction method that is particularly adept at preserving local structures and revealing clusters in high-dimensional data. By mapping the complex, multidimensional blood biomarker data onto a two-dimensional plane, t-SNE facilitates the visual inspection of the data's inherent clustering patterns. This visualization aids in understanding how well the blood biomarkers can differentiate individuals with AD from those without, essentially evaluating the biomarkers' effectiveness as diagnostic tools, as well as justifying the need for further analysis. The resulting t-SNE plot in Figure \ref{fig:tsne} highlights clusters that represent the two groups, visualizing the discriminative power of biomarkers in AD diagnosis.

Figure \ref{fig:tsne} shows a distinct separation between two main clusters but reveals closer proximity between AD and Control within each cluster, suggesting potential subgroup variations within both groups. AD clusters tend to be more concentrated than the control group with a wider spread, highlighting the complexity in using biomarkers for diagnosing AD.

\subsection{Robust Biomarker Discovery \& AD Diagnosis}
% \begin{figure*}[t!]
%     \centering
%     \includegraphics[width=\textwidth]{Figures/comb_SHAP.png}
%     \caption{Top distinguishing features for different models from SHAP analysis}
%     \label{fig:comb-shap}
% \end{figure*}
To find the biomarkers that are most important in differentiating AD from control, we utilized logistic regression (LR), random forest (RF), and shallow multi-layer perceptron (MLP) models to classify AD and control groups. Logistic regression is trained with an elastic-net penalty to encourage partial sparsity while preserving correlated biomarkers, avoiding collapse to minimal feature sets that may obscure interacting biological signals. While BRAIN theoretically accommodates any ML classification model, our selection of LR, RF, and MLP for this study is influenced by the previous works and dataset characteristics. Each model optimizes a distinct objective function or adopts a different training strategy. Evaluating multiple models together is important to avoid overlooking any biomarkers, as different models prioritize different ones.

\begin{figure*}[t]
    \centering
    \includegraphics[width=.7\textwidth]{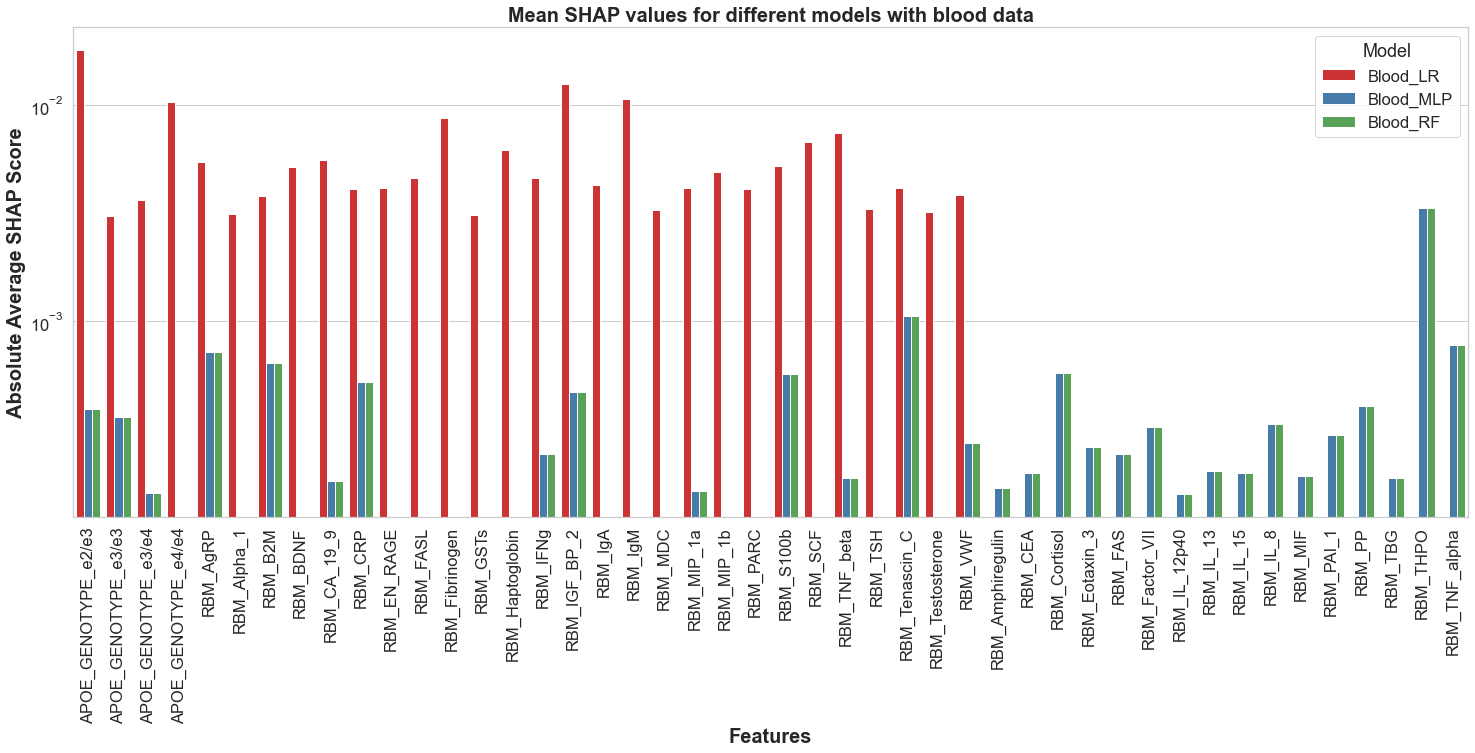}
    \caption{Top distinguishing features for different models from SHAP analysis}
    \label{fig:comb-shap}
\end{figure*}

We employ SHapley Additive exPlanations (SHAP), a method that explains model output by quantifying each feature's contribution to the model's predictions~\cite{shap}. SHAP provides additive, model-consistent, and globally comparable importance scores that can be reliably aggregated across models and bootstrap runs, which is essential for population-level biomarker discovery and network construction; local methods such as LIME are not well-suited for this objective due to their instance-specific and sampling-dependent explanations. Rooted in cooperative game theory, SHAP utilizes the concept of Shapley values, assigning a value to each feature based on its marginal contribution to the overall prediction. If a feature has no effect on the model, its SHAP value is zero. We conduct SHAP analysis on each model in a bootstrapped manner. Following model training, SHAP scores are computed for each biomarker across the test set. This process is iterated $B$ times, where in each iteration, data is randomly partitioned into training and test sets to enhance generalization. Finally, the average importance score for biomarker $i$, represented by $S_i$, is computed after $K$ iterations as,
\begin{equation}
    S_i = \frac{1}{TBK} \sum_{j=1}^{TBK} abs(s_i^j)
\end{equation}
where $T$ is the test set size, $B$ is the number of bootstrapping iterations per model, $K=3$ is the number of models, and $s_i^j$ is the individual SHAP score for biomarker $i$ and sample $j$.

We identified top $N$ biomarkers from SHAP scores for all models. Due to differences in the spread of importance scores across models, the value of $N$ varies slightly for each model based on empirical testing. Hence, we present approximately 30 top biomarkers from each model. This choice was inspired by domain knowledge and existing research on biomarker analysis for AD, ensuring that biomarkers deemed important by previous studies are included~\cite{Hampel2023}. While exact biomarkers discovered are not as pertinent at this stage, we highlight the divergence among different models regarding biomarkers, despite some overlap: biomarkers discovered by LR differ from those identified by MLP and RF, although the latter two exhibit considerable similarity. We hypothesize that this discrepancy may be due to differences in model complexity. However, one feature is notably present across models: \textit{the biomarkers corresponding to the genotype for Apolipoprotein E (APOE)} (Figure~\ref{fig:comb-shap}).

\subsubsection{Controlling for APOE as a proxy for AD}
% \begin{figure}[ht!]
%     \centering
%     \includegraphics[scale=0.3]{Figures/APOE.png}
%     \caption{Distribution of APOE Genotypes by AD. ANOVA tests assess whether there are statistically significant differences in the distribution of APOE genotypes across different AD categories.}
%     \label{fig:APOE}
% \end{figure}

% \begin{figure}[h!]
%     \centering
%     \includegraphics[scale=0.33]{Figures/IGF.png}
%     \caption{IGF-BP-2 Distribution. ANOVA confirms significant difference in biomarker for different AD classes (p-value=$1.5 x 10^{-7}$).}
%     \label{fig:IGF}
% \end{figure}

\begin{figure}[t]
    \centering
    \includegraphics[trim ={5 0 0 28}, clip, width=.7\linewidth]{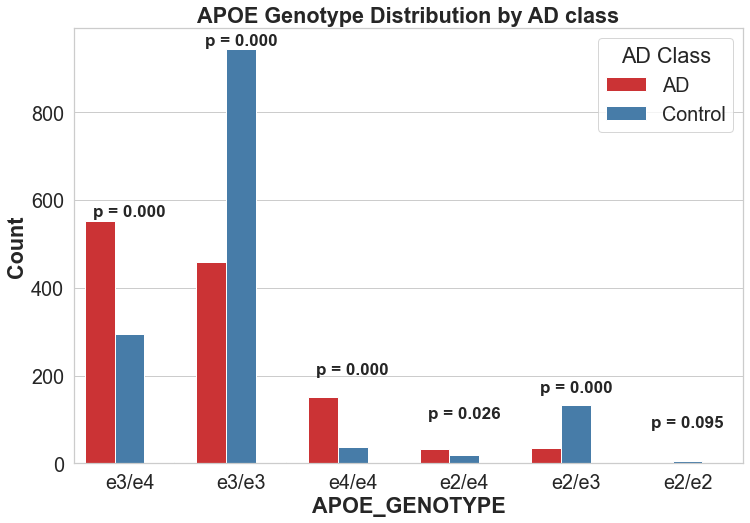}
    \caption{Distribution of APOE Genotypes; ANOVA p-values evaluate differences in APOE genotype distributions across AD and control}
    \label{fig:APOE}
\end{figure}

% \begin{figure}[ht!]
%     \centering
%     \begin{adjustbox}{minipage=0.5\textwidth,valign=b,width=0.45\textwidth,height=0.4\textwidth}
%         \centering
%         \includegraphics[width=\textwidth]{Figures/apoe_noMCI.png}
%         \caption{Distribution of APOE Genotypes; ANOVA p-values evaluate differences in APOE genotype distributions across AD and control}
%         \label{fig:APOE}
%     \end{adjustbox}
%     \hfill
%     \begin{adjustbox}{minipage=0.5\textwidth,valign=b,width=0.45\textwidth,height=0.4\textwidth}
%         \centering
%         \includegraphics[width=\textwidth]{Figures/rbm-noMCI.png}
%         \caption{IGF-BP-2 Distribution; ANOVA confirms a significant difference in biomarker value between the AD and control group (p-value < 0.01)}
%         \label{fig:IGF}
%     \end{adjustbox}
%     % \caption{Comparison of APOE Genotypes and IGF-BP-2 Distribution in AD}
%     \label{fig:combined}
% \end{figure}

APOE gene is the most prevailing risk factor of late onset AD, with the epsilon4 allele contributing a 20-25\% risk of developing AD and carrying 2 epsilon4 alleles contributing up to 50\% risk~\cite{Raulin2022}. In contrast, carriers of the epsilon2 allele are neuroprotected against AD. It is important to note that these risks vary widely by racial background~\cite{belloy_apoe_2023}. Since the link between APOE and AD is well documented, the utilization of SHAP values allowed us to screen for APOE-related biomarkers. To validate this, we plot the distribution of APOE genotypes among the AD and control groups (Figure \ref{fig:APOE}). The APOE gene is categorized by e2/e2, e2/e3, e2/e4, e3/e3, e3/e4, and e4/e4 genotypes. Certain APOE genotypes, especially those involving the e4 allele, have been associated with a higher risk of developing AD \cite{Raulin2022}. The APOE genotype distribution seems to skew towards e3 and e4 alleles within the AD group. As predicted, this suggests that these could be considered as a proxy or an indicator for AD. To further validate our hypothesis, we conducted an analysis of variance (ANOVA) test at the $\alpha = 0.05$ significance level, comparing the values of APOE genotypes between AD and control groups, where we noted statistically significant differences in the APOE genotype values among these groups.

\begin{figure}[t]
    \centering
    \includegraphics[trim ={5 0 0 28}, clip, width=.7\linewidth]{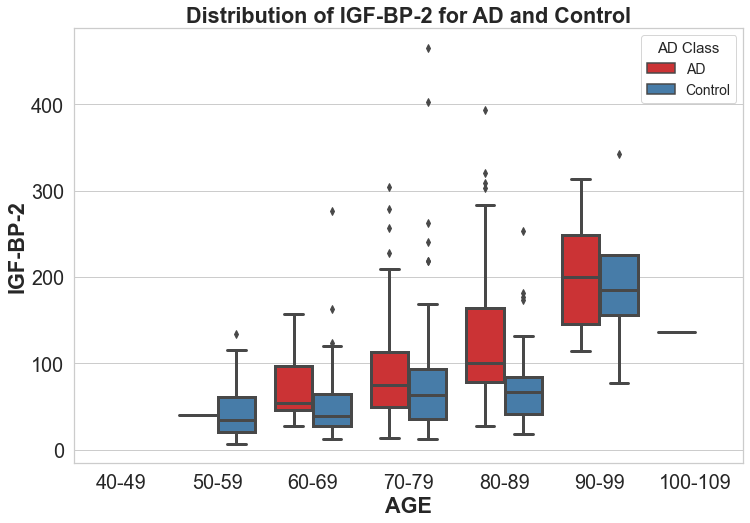}
    \caption{IGF-BP-2 Distribution; ANOVA confirms a significant difference in biomarker value between the AD and control group (p-value < 0.01)}
    \label{fig:IGF}
\end{figure}

Since APOE-related biomarkers are highly correlated with AD, they may overshadow the effects of other biomarkers in the analysis, particularly if other biomarkers are continuous variables. In line with our objectives to find new insights that may allow drug discovery, APOE genotypes were removed from further analysis. A comprehensive model encompassing APOE genotype and other biomarkers can then be communicated as part of a predictive model or for diagnostic purposes.

\subsubsection{Multicollinear biomarkers}\label{corr}

After excluding APOE-related biomarkers, we conducted regression analyses on the remaining biomarkers to evaluate their impact on the likelihood of disease occurrence, with statistical significance determined by p-values. This analysis was conducted to understand how the significance of specific biomarkers depends on a particular combination of biomarkers incorporated in the model.

We refrain from presenting the whole analysis due to space limitations but provide one example from our findings to illustrate the issue of multicollinearity in this complex problem. The biomarker for insulin-like growth factor-binding protein 2 (IGF-BP-2) was initially not statistically significant when differentiating AD vs control (p>0.05). However, when considered independently using an ANOVA between IGF-BP-2 differences in age groups, IGF-BP-2 was found to be significant at the $\alpha$ = 0.05 significance level. Figure \ref{fig:IGF} illustrates an increase in IGF-BP-2 levels associated with aging. When the model was adjusted to account for age, it was not statistically significant anymore. Importantly, we noted that the differences in IGF-BP-2 levels between the AD and control groups became more marked with age, indicating a potential interaction between age and disease progression in influencing this biomarker's levels, a finding that was recently confirmed in a large scale healthcare study on IGF-BP-2 \cite{McGrath2019}. 

\begin{figure*}[ht!]
    \centering
    \begin{subfigure}[b]{0.45\textwidth}
    \centering
        \includegraphics[width=.8\textwidth]{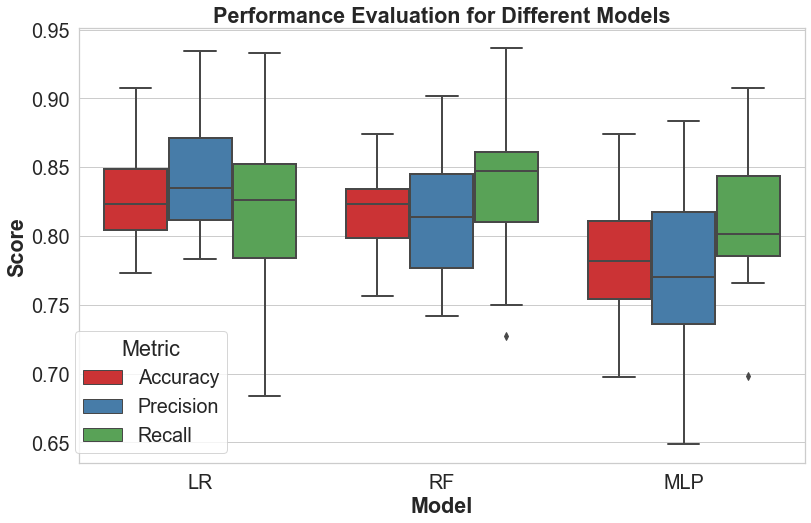}
        \caption{Performance of different models for AD prediction (multiple runs of the model evaluation experiment using bootstrapping, which provides a robust estimate of model performance)}
        \label{fig:AD-pred_modles}
    \end{subfigure}
    \hfill
    \begin{subfigure}[b]{0.45\textwidth}
    \centering
        \includegraphics[width=.9\textwidth]{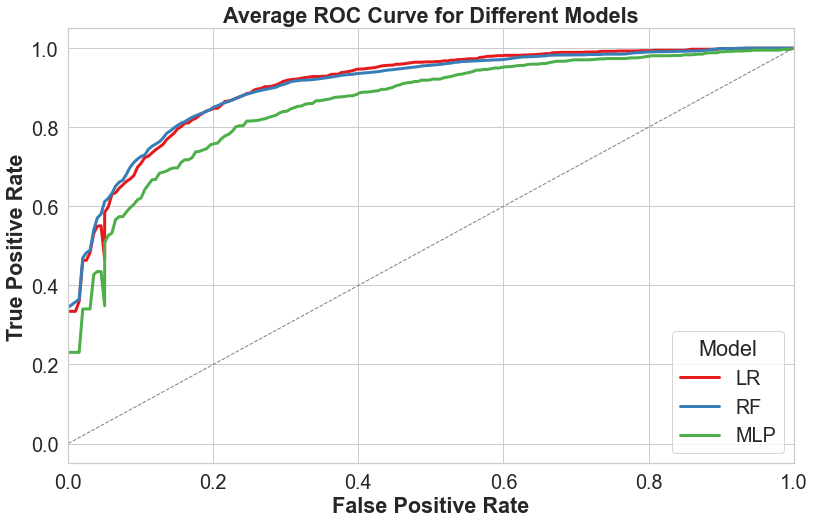}
        \caption{ROC curves for different models}
        \label{fig:roc}
    \end{subfigure}
    \caption{Performance of ML models for AD classification}
\end{figure*}

% \begin{figure}[ht!]
%     \centering
%     \includegraphics[scale=0.25]{Figures/IGF.png}
%     \caption{IGF-BP-2 distribution with age group changes statistically significantly between AD and control.}
%     \label{fig:IGF}
% \end{figure}

This analysis shows how models with similar performance may assign high importance to different sets of biomarkers. It highlights the role of inter-biomarker correlations, which can confound the discovery process. Specifically, the correlation between biomarkers may lead to issues such as multicollinearity, where predictors are not independent of one another. This can affect the stability and interpretability of the model coefficients, potentially leading to different models prioritizing different biomarkers despite achieving comparable levels of predictive accuracy, further underscoring a need for understanding their inter-dependencies from a more "holistic" lens. 

The implications of these findings are important for statistical analysis, which is commonly used in literature for biomarker discovery. Correlated biomarkers may not provide independent information about the disease state, and their individual significance could be overestimated unless the correlation is accounted for. Hence, in hypothesis testing, such as when determining if a particular biomarker is significantly associated with AD, it is important to include other correlated biomarkers in the model to control for their shared variance. In the clinical context, understanding biomarker interdependencies is crucial for utilizing them effectively. If two biomarkers provide similar information due to high correlation, measuring only one could offer a more streamlined and cost-efficient approach. This is particularly relevant in personalized medicine, where such knowledge can help customize treatment plans to an individual's unique disease profile, enhancing the efficacy of interventions. Furthermore, in research involving high-dimensional data, recognizing biomarker correlations is key to mitigating the risk of false discoveries that can arise from multiple tests. Incorporating these insights into predictive models can also enhance their precision, leading to better tools for prognosis.

\section{Experiments \& Results}

For each model-data segment combination, the data was split into a 80\%-20\% train-test split and a 5-fold cross validation was carried out for hyperparameter tuning using grid search over the hyperparameter space. To ensure replicability, bootstrapping was conducted where this entire process was repeated 20 times, with a different train-test split each time. We utilized bootstrapping and feature importance interpretation for model evaluation. Bootstrapping allowed us to generate a distribution of results, particularly micro-F1 score, sensitivity, and specificity. Complementary with confusion matrices, these metrics best capture classification of true positives and true negatives, allowing us to compare our performance with similar models. The experiments were conducted on a Macbook with an Apple M2 Pro chip, 32GB Memory, and 2TB Storage. The entire experiment, from data cleaning to final output networks, took less than 30 minutes.

\subsection{Alzheimer's Disease Classification Results} \label{model_perf}

\begin{table}[t]
\centering
\caption{ Confusion matrix counts are aggregated across bootstrap iterations, while recall (TPR) and specificity (TNR) are reported as mean $\pm$ standard deviation across bootstraps.}
\label{tab:blood_counts_rates}
\begin{tabular}{lcccccc}
\toprule
Model & TN & FP & FN & TP & TPR (Recall) & TNR (Spec.) \\
\midrule
LR  & 976 & 185 & 223 & 996 & $0.818 \pm 0.056$ & $0.840 \pm 0.045$ \\
RF  & 958 & 226 & 200 & 996 & $0.834 \pm 0.051$ & $0.810 \pm 0.043$ \\
MLP & 889 & 288 & 228 & 975 & $0.810 \pm 0.050$ & $0.757 \pm 0.067$ \\
\bottomrule
\end{tabular}
\end{table}

We evaluated three ML models -- LR, RF, and Multi-Layer Perceptron (MLP) -- for AD prediction using blood biomarker data. The elastic-net logistic regression model used regularization values spanning several orders of magnitude with parameters in the range $[0, 1]$, while the MLP employed two hidden layers of 20 units each and was trained for a maximum of 300 epochs. As shown in Figure~\ref{fig:AD-pred_modles}, LR exhibits the lowest median accuracy with a broader interquartile range, indicating variability and generally lower performance. In contrast, RF and MLP achieve higher accuracy, with MLP slightly outperforming RF. LR again shows lower precision, while MLP demonstrates a marginal advantage over RF. LR lags behind in recall with lower median values and greater spread, whereas RF and MLP are more consistent, with MLP displaying minor superiority.

\begin{figure}[htb!]
    \centering
    \includegraphics[trim ={57 25 57 25}, clip, width=.7\linewidth]{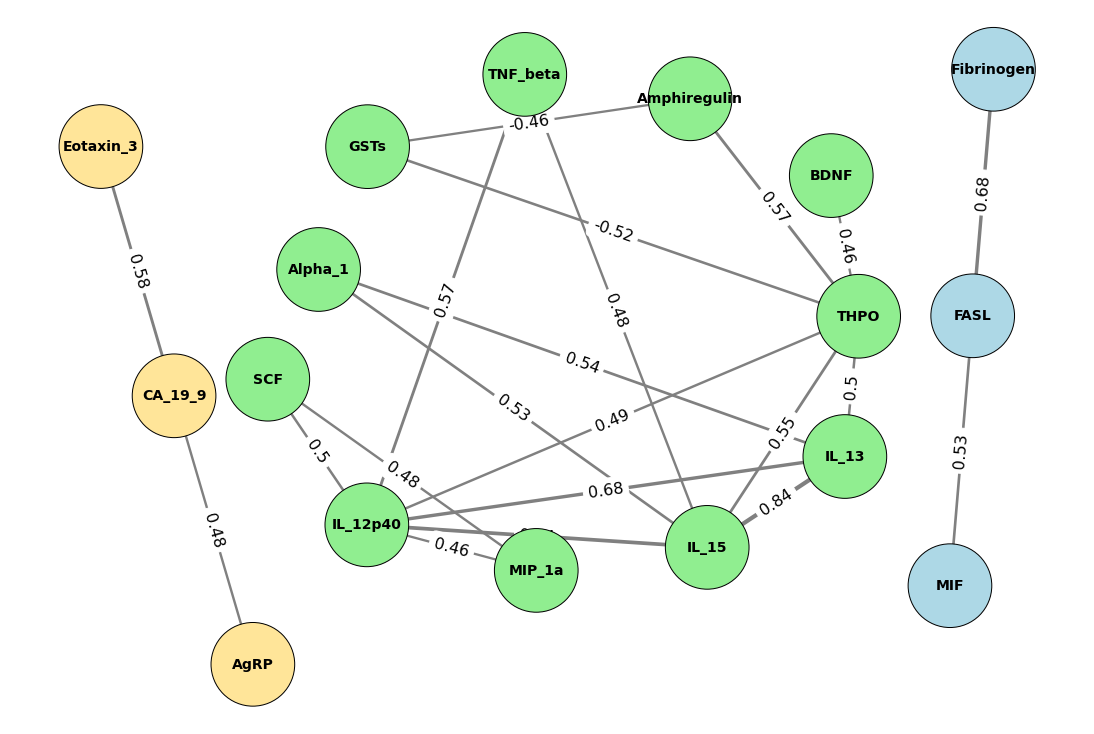}
    \caption{Graph structure for all data (AD and control), $\alpha$ = 0.45}
    \label{fig:comb_graph}
\end{figure}

ROC curves (Figure~\ref{fig:roc}) reinforce these observations, demonstrating that MLP achieves a higher true positive rate for blood biomarker data. All three models significantly outperform random classification (diagonal line), with MLP showing the highest area under the curve (AUC), followed by RF and LR.

To further characterize model error patterns beyond scalar performance metrics, Table~\ref{tab:blood_counts_rates} reports aggregated confusion matrix counts along with recall (TPR) and specificity (TNR) summarized as mean $\pm$ standard deviation across bootstrap iterations. Consistent with Figures~\ref{fig:AD-pred_modles} and~\ref{fig:roc}, recall remains comparable across models, while MLP exhibits lower specificity and higher variability, reflecting a higher false-positive rate. In contrast, LR and RF achieve more balanced trade-offs between sensitivity and specificity, with RF showing the most stable recall across resampling.

\subsection{Graph network representation reveals holistic  interrelations} \label{graphs-section}
Analyzing biomarker correlation matrices poses significant challenges, primarily due to complex interrelations among biomarkers, hindering pattern isolation and complicating visualization. These difficulties are exacerbated with increasing data dimensionality, amplifying the difficulties associated with information extraction and interpretation of these heatmaps. \\
To enhance explainability, we adopt a novel graph network approach where we utilize graph representation to showcase the relationships between different biomarkers. To this end, we define a graph $\mathcal{G}=(V,E)$, where $V$ represents nodes and $E$ are edges between them. The nodes in this network are biomarkers, and edges represent the relationship between them. This relationship is proportional to the correlation between the biomarkers. The graph is constructed between the pool of biomarkers identified by SHAP analysis without the APOE genotypes. For each distinct biomarker pair  $B_i[k] , B_j[k]$ where $i \neq j$ and sample index $k$, over a population size $n$, we generate the weight $w_{i,j}$ of edge $E_{i,j}$,
\begin{subequations}
\begin{align}
& w_{i,j} = \frac{(\sum_{k=1}^{n} B_{i}[k] B_{j}[k]) - (\frac{1}{n}\sum_{k=1}^{n} B_{i}[k])(\sum_{k=1}^{n} B_{j}[k])}
{\sqrt{\text{Var}(B_i) \cdot \text{Var}(B_j)}} \\
& \text{where,} \quad \text{Var}(B_i) = \frac{1}{n}[n\sum_{k=1}^{n} B_{i}[k]^2 - (\sum_{k=1}^{n} B_{i}[k])^2] 
\end{align}
\end{subequations}

To detect meaningful patterns in the correlation matrix,  differentiate between spurious correlations and genuine relationships, and create better visualizations for clearer interpretation, we use a threshold $\alpha$ to drop edges below a certain weight/correlation,
\begin{equation}
E_{i,j} = \begin{cases} 
0 & \text{if } w_{i,j} < \alpha \\
w_{i,j} & \text{otherwise} 
\end{cases}
\end{equation}
% \begin{figure}[htb!]
%     \centering
%     \includegraphics[scale=0.2]{Figures/SHAP_Full_graph.png}
%     \caption{Graph structure for all data (AD and Control), $\alpha$=0.45}
%     \label{fig:comb_graph}
% \end{figure}

% \begin{figure}[htb!]
%     \centering
%     \begin{subfigure}[b]{0.49\textwidth}
%         \includegraphics[width=1.1\textwidth]{Figures/SHAP_AD_graph.png}
%         \caption{Graph network for AD}
%         \label{fig:sub1}
%     \end{subfigure}
%     \hfill 
%     \begin{subfigure}[b]{0.49\textwidth}
%         \includegraphics[width=1.1\textwidth]{Figures/SHAP_Control_graph.png}
%         \caption{Graph network for  Control}
%         \label{fig:sub2}
%     \end{subfigure}
%     \caption{Graph Networks for AD and Control ( $\alpha=0.45$)}
%     \label{fig:class_wise_graph}
% \end{figure}
% \begin{figure}[t!]
%     \centering
%     \includegraphics[scale=0.32]{Figures/Degree_dist.png}
%     \caption{Degree distribution in different networks}
%     \label{fig:deg_dis_graph}
% \end{figure}

\begin{figure*}[t]
    \centering
    \begin{subfigure}[b]{0.49\textwidth}
        \centering
        \includegraphics[trim ={57 25 57 25}, clip, width=.7\linewidth]{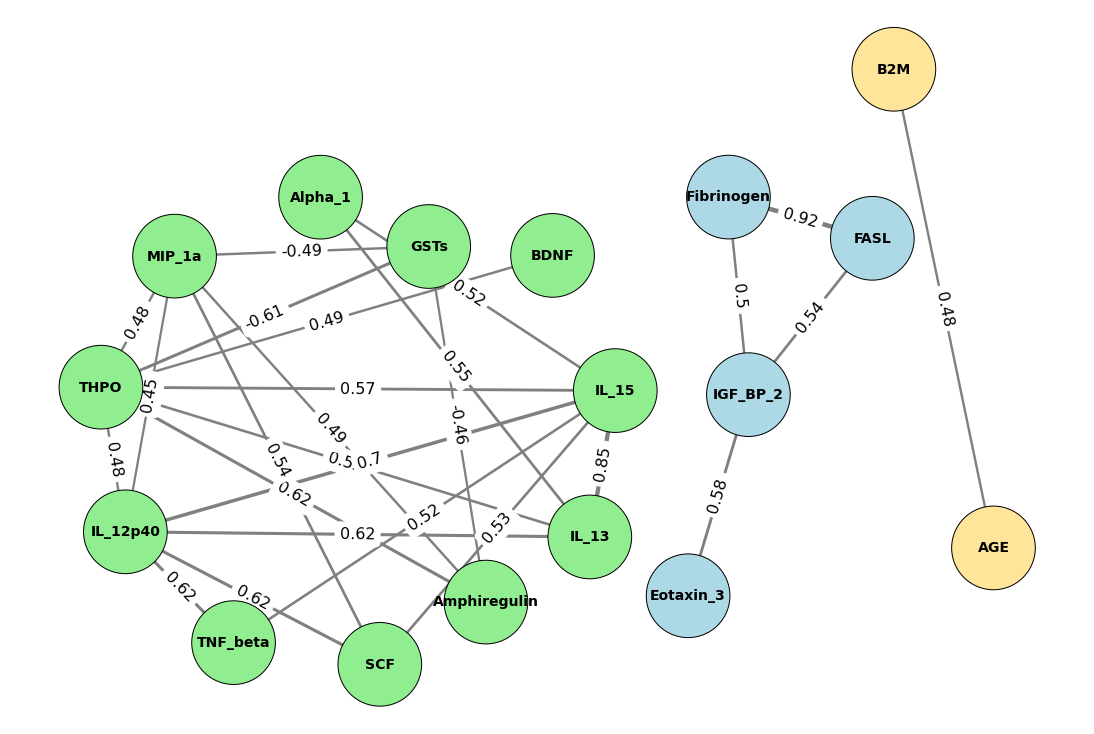}
        \caption{Graph network for AD}
        \label{fig:sub1}
    \end{subfigure}
    \begin{subfigure}[b]{0.49\textwidth}
        \centering
        \includegraphics[trim ={57 25 57 25}, clip, width=.7\linewidth]{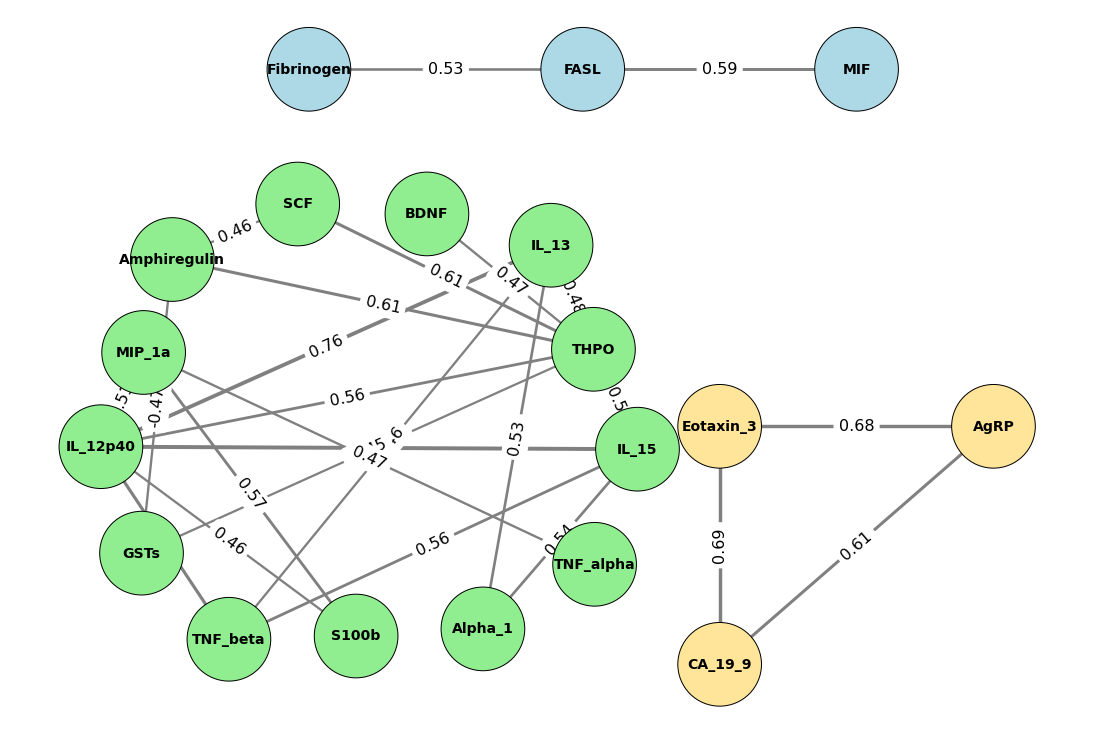}
        \caption{Graph network for  Control}
        \label{fig:sub2}
    \end{subfigure}
    \caption{Graph networks for AD and control ($\alpha=0.45$)}
    \label{fig:class_wise_graph}
\end{figure*}

Finally, we visualize the graph for biomarkers that have any interconnections (Figure~\ref{fig:comb_graph}), with $\alpha = 0.45$. If a biomarker has no edge, we do not include it in the visualization to minimize clutter and improve visual clarity. The threshold, $\alpha$, was chosen based on empirical testing and domain knowledge, ensuring that biomarkers that have been identified as important contributors in AD are included in the graph~\cite{Hampel2023}. The edge weights are presented in the edges and the nodes represent biomarkers. We find three components/clusters within the graph. Each component is colored differently to highlight distinct groups. Looking closer, the graph shows that the biomarkers Exotaxin-3, CA-19-9 and AgRP form one group as CA-19-9 is highly correlated with the other two. The second group  has a similar trend with the biomarker FASL having a connection to Fibrinogen and MIF. The third component in green is the most complicated and dense with 11 members. In this group, biomarkers THPO and IL-12p40 are highest connected nodes.

Next, we analyze the impact of AD on these interconnections by separating the data into AD and control groups and repeating the graph construction process described above with the same threshold parameter $\alpha$. The networks extracted for AD and control are shown in Figure~\ref{fig:sub1} and Figure~\ref{fig:sub2}, respectively. Again, we observe 3 components in each network; however, the size of the component and individual biomarkers are different. Comparing blue components in both graphs, we observe that Fibrogen-FASL link is part of different networks in each graph and their strength is also significantly stronger in AD network. This link is connected to IGF-BP-2-Eotaxin-3 in AD graph compared to connection with MIF in control graph. Eotaxin in control graph is linked to CA-19-9 and AgRP similar to what we saw in Figure \ref{fig:comb_graph}. The complex green component still holds a large number of biomarkers with high interconnections. The size of this complex group is 11 and 13 in AD and control group respectively where TNF-Alpha and S100b present in control no longer show up in AD group.

\begin{figure}[t]
    \centering
    \includegraphics[trim ={0 0 0 28}, clip, width=.7\linewidth]{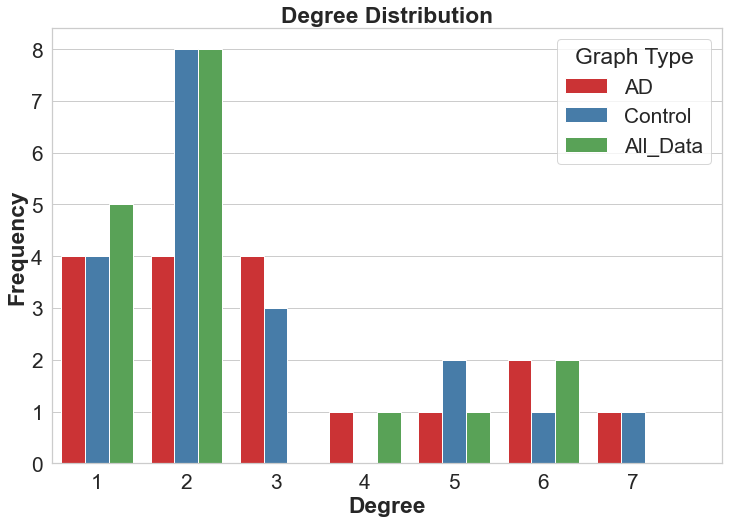}
    \caption{Degree distribution in different networks}
    \label{fig:deg_dis_graph}
\end{figure}

We systematically compare the three networks extracted from biomarker data through degree, i.e., the number of direct neighbors each node has in a graph. First, we compare the distribution of degree in the three networks in Figure~\ref{fig:deg_dis_graph}. It can be observed that the graph topologies are varying for different groups, thus indicating that overall structure is helpful in distinguishing between AD and control. Moreover, this difference becomes more prominent with increasing degree signifying the use of holistic and sophisticated features in AD detection. 
To further investigate how each biomarker changes its network topology in these networks, we present the degree of all biomarkers in Table~\ref{tab:deg-table}. If a biomarker is absent from a graph, its degree is considered zero. It can be observed that 3 distinct biomarkers appear significant in the AD network.

\begin{table}[t]
    \caption{Degree distribution of important biomarkers across different networks}
    \centering
    \begin{tabular}{lccc}
        \hline
        \textbf{Biomarker} & \textbf{\begin{tabular}[c]{@{}c@{}}AD\\Degree\end{tabular}} & \textbf{\begin{tabular}[c]{@{}c@{}}Control \\Degree\end{tabular}} & \textbf{\begin{tabular}[c]{@{}c@{}}Present Only in AD \\ (True/False)\end{tabular}} \\ \hline
        \textbf{Age at Diagnosis} & \textbf{1 }& \textbf{0} & \textbf{True} \\ \hline
        \textbf{B2M} &\textbf{1} &\textbf{ 0} & \textbf{True} \\ \hline
        $\alpha$-1 & 2 & 2 & False \\ \hline
        IL-13 & 4 & 5 & False \\ \hline
        IL-15 & 6 & 5 & False \\ \hline
        BDNF & 1 & 1 & False \\ \hline
        THPO & 7 & 7 & False \\ \hline
        FASL & 2 & 2 & False \\ \hline
        Fibrinogen & 2 & 1 & False \\ \hline
        \textbf{IGF-BP-2} &\textbf{ 3} &\textbf{ 0} &\textbf{ True }\\ \hline
        GSTs & 3 & 2 & False \\ \hline
        MIP-1$\alpha$ & 5 & 3 & False \\ \hline
        Amphiregulin & 3 & 3 & False \\ \hline
        Eotaxin-3 & 1 & 2 & False \\ \hline
        SCF & 3 & 2 & False \\ \hline
        IL-12p40 & 6 & 6 & False \\ \hline
        TNF-$\beta$ & 2 & 3 & False \\ \hline
        AgRP & 0 & 2 & False \\ \hline
        CA 19-9 & 0 & 2 & False \\ \hline
        MIF & 0 & 1 & False \\ \hline
        S100b & 0 & 2 & False \\ \hline
        TNF-$\alpha$ & 0 & 1 & False \\ \hline
        \end{tabular}%
    \label{tab:deg-table}
\end{table}

To assess the robustness of the inferred biomarker interaction graph, we conducted additional sensitivity and stability analyses. First, we evaluated network construction across a range of correlation thresholds $(0.40-0.50)$ and observed substantial edge overlap with the reference graph (Jaccard similarity $\approx$ 0.57-0.64), indicating that the overall structure is not threshold-specific. Second, using bootstrap-based stability selection (200 resamples), we identified a core set of 14 edges that appeared in at least 70\% of resamples, with several key associations exhibiting near-perfect stability. Finally, we performed a partial-correlation analysis using a shrinkage precision estimator, which yielded qualitatively consistent core connections, suggesting that the identified modules are not driven solely by indirect effects. Together, these analyses support the robustness and reliability of the inferred biomarker modules.

\section{Discussion}
This study introduces a novel methodology for investigating AD biomarkers by prioritizing network-level holistic indicators over isolated biomarkers. Because many blood biomarkers are correlated and biologically interdependent, purely sparse predictive models may suppress relevant features; our objective is therefore to identify stable, interaction-aware biomarker modules rather than minimal sparse predictor sets. As depicted in Figure \ref{fig:comb_graph}, the network-level interactions between the control and AD networks undergo significant changes, with certain inflammatory pathway biomarkers—namely, IL-13, IL-15, MIP-1$\alpha$, and IL-12p40—demonstrating heightened edge scores, indicative of an increased correlation within the AD network. This supports the theory that inflammatory pathways play a crucial role in AD progression and emphasizes the importance of targeting these pathways for therapeutic development.

Accounting for age, our analysis enables age-agnostic interpretations. Notably, IGF-BP-2 and B2M, absent in the control network, emerge within the AD network, suggesting potential associations with Alzheimer's pathology. Increased expression of IGF-BP-2 is linked to AD progression in asymptomatic individuals \cite{quesnel_insulin-like_2023}, while B2M is associated with cerebrospinal fluid (CSF) AD biomarkers, $\beta$-amyloid pathology, and cognitive impairment in individuals who are cognitively normal or in preclinical stages of AD \cite{huang_plasma_2023}. These findings corroborate existing research on early-stage AD biomarkers and validate the efficacy of graph network analysis in biomarker exploration. 

Several biomarkers prevalent in late-stage AD do not exhibit consistent degrees across all groups, indicating their limited utility in distinguishing AD from control cases, e.g., Alpha-1, associated with late-stage AD lesions~\cite{gollin_alpha_1992}, and glutathione S-transferases (GSTs), linked to increased AD risk in the elderly~\cite{allen_glutathione_2012}. Although IL-12p40 and THPO are closely correlated with other inflammatory biomarkers, their predictive value for differentiating AD is limited. IL-12p40's link to inflammatory load in AD~\cite{pedrini_blood-based_2017} and THPO's association with the progression from mild cognitive impairment to dementia~\cite{royall_thrombopoietin_2016} imply that despite their correlations with AD and control networks, these biomarkers are independently insufficient to determine AD.

We note certain limitations in our data and methods. The TARCC dataset exhibits sampling bias and is relatively small in size, potentially limiting the generalizability of the results. We attempt to compensate for this gap through validation with biomedical domain knowledge. Furthermore, unlike prior AD research on high-dimensional imaging data, we employ models of limited complexity due to dataset size constraints, preventing the utilization of more sophisticated DL models. Scaling up the dataset or exploring techniques for training DL models on small datasets could address this limitation. In addition, our graph construction uses Pearson correlation to define edge weights, which captures linear associations rather than causal relationships and may not capture non-linear or higher-order interactions. The graph representation is designed to visualize biomarker interdependencies and inform hypothesis generation for future mechanistic studies that could employ causal inference methods. Additionally, categorical or binary biomarkers cannot be incorporated in the graph representation due to constraints on how edge weights are defined since the vast majority of blood-based biomarkers in the scope of such an analysis are continuous variables.

Currently, we choose design parameters, such as the number of features filtered through SHAP and the threshold, $\alpha$, based on domain knowledge via existing biomedical literature. In future, this process of parameter selection can be automated. Additionally, BRAIN employs multiple models that could pose computational resource issues, especially with increasingly large datasets. However, current blood-based datasets do not reach extreme sizes due to the high cost of human data collection. As technology advances and larger datasets become more accessible, computational challenges may become more pronounced and require innovative solutions. Finally, we employ SHAP, which assumes additivity in the relationship between model output and features, an assumption that may not always hold. Future work could explore other feature importance techniques, such as LIME, to broaden the pool of important biomarkers and potentially uncover non-additive relationships between features and outcomes. 

% \subsection{Impact}
Prior work in biomarker-based diagnosis of AD has been exceedingly "performance" driven, i.e. solely focusing on maximizing one or more accuracy metrics in AD diagnosis. However, many state-of-the-art modeling methodologies are inherently non-interpretable or offer low biomedically-meaningful interpretability. While high AD diagnostic accuracy using biomarkers is a baseline objective in this paper, our emphasis has been to examine \textit{(i) how those biomarkers are associated with one another, and (ii) what biomedically-meaningful insights we can glean from how their correlations change from AD to control}. BRAIN addresses criticisms related to stability and reliability of interpretable models by aggregating findings from multiple models. It enables the discovery of distinct biomarker clusters in control and AD groups, highlighting significant differences in their network structures. The framework's ability to visualize and interpret complex biomarker networks provides novel insights into AD's molecular interactions and pathology. These insights are crucial for developing targeted drug therapies and enhancing early AD diagnosis, potentially transforming the treatment and management of AD. Finally, we also attempt to make ML more accessible to clinicians and biomedical researchers who are not ML experts. By providing clear visualizations and interpretable results via graphs, it arms clinicians with clear and potentially actionable insights that are biomedically meaningful.

\section{Conclusion}
We propose a new framework for investigating biomarkers for AD diagnosis and therapeutic treatment through BRAIN. When biomarkers are correlated, it indicates that they are part of a complex biological network, where changes in one could influence or be indicative of changes in another. This complexity is more than a statistical challenge; it reflects the intricate biological processes that may lead to the onset and progression of AD. We utilize an ensemble approach to find a comprehensive set of important biomarkers and, for interpretability, utilize graph networks to represent them while highlighting their interconnectedness. Finally, we analyze how these networks differ between AD and control groups. Accurate interpretation of correlations is vital for precise hypothesis testing and avoiding multicollinearity, which can obscure the impact of individual biomarkers. Inter-biomarker relationships can guide the discovery of biomarker-based therapeutics, revealing candidates for drug target investigations that may have been overlooked. These correlations offer a window into the biomedical pathways of AD, helping to chart the sequence of biological events and identify potential intervention points.

\bibliographystyle{IEEEtran}
\bibliography{iclr2023_conference}
\end{document}